\documentclass{article}
\usepackage{spconf,amsmath,graphicx}
\usepackage{float}
\usepackage{url}

\title{Redefining Cystoscopy with AI: Bladder Cancer Diagnosis using an Efficient Hybrid CNN-Transformer Model}

\name{%
    \parbox{\linewidth}{\centering
      Meryem Amaouche$^{\star }$ \qquad Ouassim Karrakchou$^{\star }$ \qquad Mounir Ghogho$^{\star }$ $^{\ddagger}$ \\ \textit{Anouar El Ghazzaly}$^{\dagger}$ \qquad \textit{Mohamed Alami}$^{\dagger}$ \qquad \textit{Ahmed Ameur}$^{\dagger}$
    }%
  }%

\address{$^{\star }$ International University of Rabat, TICLab, Morocco \\
$^{\ddagger}$ University of Leeds, Faculty of Engineering, UK\\
$^{\dagger}$ Mohammed V Military Training Hospital, Rabat, Morocco}
\begin{document}
%
\maketitle
\begin{abstract}
Bladder cancer ranks within the top 10 most diagnosed cancers worldwide and is among the most expensive cancers to treat due to the high recurrence rates which require lifetime follow-ups. The primary tool for diagnosis is cystoscopy, which heavily relies on doctors’ expertise and interpretation. Therefore, annually, numerous cases are either undiagnosed or misdiagnosed and treated as urinary infections.  To address this, we suggest a deep learning approach for bladder cancer detection and segmentation which combines CNNs with a  lightweight positional-encoding-free transformer and dual attention gates that fuse self and spatial attention for feature enhancement. The architecture suggested in this paper is efficient making it suitable for medical scenarios that require real time inference. Experiments have proven that this model addresses the critical need for a balance between computational efficiency and diagnostic accuracy in cystoscopic imaging as despite its small size it rivals large models in performance.
\end{abstract}
\section{Introduction}
\label{sec:intro}

Bladder cancer is the 10th most diagnosed cancer worldwide, with approximately 600,000 new cases and over 200,000 deaths annually  \cite{WHOBladderCancer}. Cystoscopy, the primary diagnostic technique, is a procedure that involves the use of a Cystoscope in order to allow the doctor to see and examine the internal surface of the bladder. This procedure is operator dependent as it directly relies on the urologist's skills and expertise \cite{chai2021comparing}. As a result, the interpretations can vary from one doctor to another. Furthermore, estimates suggest that white-light cystoscopy fails to detect $10\%$ to $20\%$ of bladder tumors \cite{kausch2010photodynamic}. 

The rise of AI in medical imaging has introduced new computer-aided techniques to enhance diagnostic accuracy.  However, the deployment of these models in medical scenarios which require real time inference is limited by their high computational complexity. This gap requires a model that balances efficiency with accuracy.
Many studies performed classification on Cystoscopy images \cite{ikeda2020support} \cite{lorencin2020using}. Our research, however, extends to semantic segmentation, aiming to not only detect tumors but also map their covered area and exact location, crucial for real-time diagnostic guidance.
Since the introduction of Fully Convolutional Networks (FCNs \cite{long2015fully}), they have become the backbone for most semantic segmentation tasks. However, due to the nature of convolutional operations, FCNs are primarily limited by their focus on local features and inability to capture global contextual information effectively. Consequently, recent advancements introduced Transformers, known for excelling in large-scale global context interpretation. However, they require large datasets for training and struggle with high-resolution images due to the quadratic complexity of self-attention computation. Transformers also tend to face challenges in capturing fine-grained local details due to the lack of inductive biases inherent in CNNs. 
Given these complementary strengths, our study explores a hybrid  CNN-Transformer approach, aiming to leverage the global information modeling of transformers and the fine spatial detail extraction of CNNs while maintaining a lightweight architecture. This combination offers a promising solution for efficient and precise medical image segmentation, addressing the inherent limitations of each individual approach. It is worth noting that hybrid models such as TransUNet \cite{chen2021transunet} and UNETR  \cite{hatamizadeh2022unetr}  have shown impressive results in medical imaging. 

Last but not least, a significant limitation in developing a bladder cancer segmentation model is the lack of annotated datasets and limited access to cystoscopy recordings due to medical confidentiality. For this reason, we collaborated with the Urology Department of the Mohammed V Military Hospital to create a cystoscopy segmentation dataset.
The main contributions of this paper can be described as follows:
\begin{itemize}
\item In collaboration with medical experts, we have compiled and annotated a comprehensive cystoscopy dataset tailored for bladder segmentation.  We ensured dataset diversity by integrating images obtained through a variety of cystoscopes, while also encompassing a broad spectrum of tumor presentations observed across different patient cases.
\item we have developed a lightweight hybrid semantic segmentation model that combines CNNs with an efficient positional-encoding-free transformer and dual attention gates for improved feature fusion in skip connections. Despite its compact size, this model is shown to deliver competitive results compared to much larger models, demonstrating an excellent balance between accuracy and computational efficiency. This makes it ideal for real-time bladder cancer diagnosis.
\end{itemize}

\section{Related work}
\label{sec:rel}
\textbf{Bladder cancer detection}
The need for precise and early bladder cancer detection has led to the adoption of deep learning techniques. Ikeda et al. \cite{ikeda2021cystoscopic} used a pre-trained GoogLeNet for tumor classification, while Ali et al. \cite{ali2021deep} utilized various architectures, including InceptionV3, MobileNetV2, ResNet50, and VGG16, on blue-light cystoscopy images. Lazo et al. \cite{lazo2023semi}, used Cycle-GAN to address label scarcity.
In semantic segmentation and object detection, S. Wu et al. \cite{wu2022artificial} leveraged a pre-trained PSPNet, while Yoo et al. \cite{yoo2022deep} applied a Mask RCNN with a ResNeXt-101-32×8d-FPN backbone, achieving a Dice Coefficient (DC) of $74.7\%$. Mutaguchi et al. \cite{mutaguchi2022artificial} demonstrated Dilated UNet's superiority over traditional UNet with a DC of $83\%$. Additionally, Zhang, Qi, et al. \cite{zhang2022attention} proposed an attention-based model, achieving a DC of $82.7\%$, while \cite{negassi20203d} obtained a DC of $67\%$.\\
\textbf{CNN vs Transformer based semantic segmentation}
The introduction of FCN  \cite{long2015fully}  sparked a revolution in semantic segmentation, leading to the emergence of many new techniques like DeepLab \cite{chen2017deeplab} and PSPNet \cite{zhao2017pyramid}, which introduced atrous convolutions and spatial pyramid pooling for contextual enhancement. Attention-UNet \cite{oktay2018attention} incorporated attention to refine feature selection,, while DFANet \cite{li2019dfanet} and BiseNet \cite{yu2018bisenet} aimed for efficiency with lightweight architectures.\\
In contrast, transformers, originally used in natural language processing, excel in global context modeling. Transformers, such as ViT \cite{dosovitskiy2020image}, have applied this concept to image recognition by treating images as 2D patches, similarly to words in NLP tasks. However, despite efforts to create more efficient transformer architectures, models like DETR \cite{zhu2020deformable}, Swin Transformer \cite{liu2021swin}, and Segformer \cite{xie2021segformer} can still face computational challenges with large images and may struggle with local detail representation on small datasets. Our research combines the strengths of CNNs, known for fine spatial detail capture, with efficient transformers to address these limitations, offering a promising solution for bladder cancer segmentation without the need for extensive datasets.
\textbf{Non-local processing in computer vision}
The self attention mechanism used in transformers can be seen as a special case of the non-local mean \cite{buades2005non}, which is a filtering algorithm in computer vision. This approach inspired the development of Non-local Neural Networks \cite{wang2018non}, a deep learning architecture designed to capture long-range dependencies in data, regardless of their spatial or temporal separation. This technique has proven to be useful for tasks where long-range dependencies are crucial, such as in video classification, object detection, and semantic segmentation.
\section{PROPOSED METHOD}
\label{sec:met}
\subsection{Network Architecture}
\label{ssec:Arch}
\begin{figure*}[]
\centering
\includegraphics[width=\textwidth]{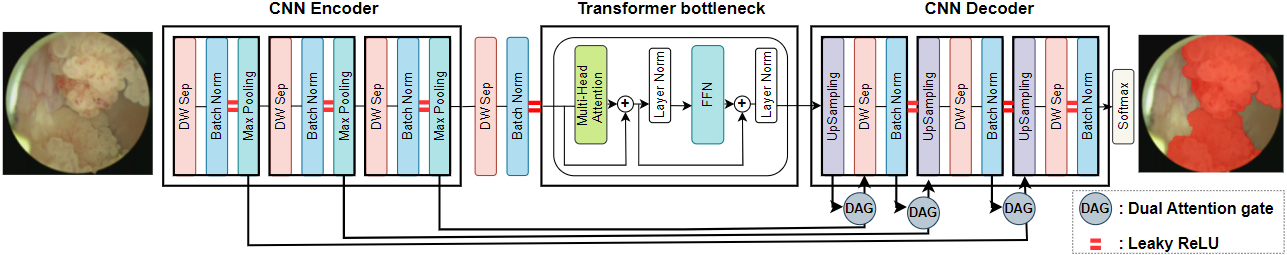}
\caption{Architecture of the Lightweight CNN-Transformer Network for bladder cancer segmentation.}
\label{fig:architecture}
\end{figure*}

The proposed architecture, illustrated in Fig. \ref{fig:architecture}, is a U-shaped structure that includes an encoder, decoder, transformer bottleneck, and dual attention gates. It processes 2D Cystoscopy images using a CNN encoder to capture fine local details and short-range dependencies. These features are then fed into an efficient transformer block, which models complex spatial information and long-range feature dependencies. Building upon the concept of attention gates (AGs) \cite{oktay2018attention}, we introduce dual attention gates that incorporate self-attention within the skip connections through the use of two distinct attention paths: self-attention and spatial attention. This enhancement captures long-range dependencies across feature maps while preserving fine-grained local features, outperforming traditional AGs.
For optimal efficiency, we used Depth-Wise (DW) \cite{chollet2017xception} separable convolutions all along the encoder and the decoder blocks. This approach significantly lowers the number of parameters and computational complexity. 
\subsection{Dual Attention Gates (DAGs)}
\label{ssec:Att}
Skip connections are a crucial part of UNet. However, the concatenation process can bring along poor feature representations from initial layers. To address this, Oktay et al. \cite{oktay2018attention} introduced attention gates (AGs) within skip connections. These gates employ a form of spatial attention to selectively focus the model on relevant regions by adaptively weighting the feature maps. However, this technique lacks the ability to grasp global relationships between different features in the feature map, which is crucial in medical imaging where certain tumors, could span a wide area in the image.
To address this, we used a dual-attention strategy  by integrating self attention into the attention gates alongside the spatial attention. The self-attention mechanism is implemented by incorporating the principle of non-local networks \cite{wang2018non}. This dual attention approach maintains focus on significant regions through spatial attention while also capturing long-range spatial dependencies with self-attention.\\
\begin{figure}[]
    \centering
    \includegraphics[width=6cm]{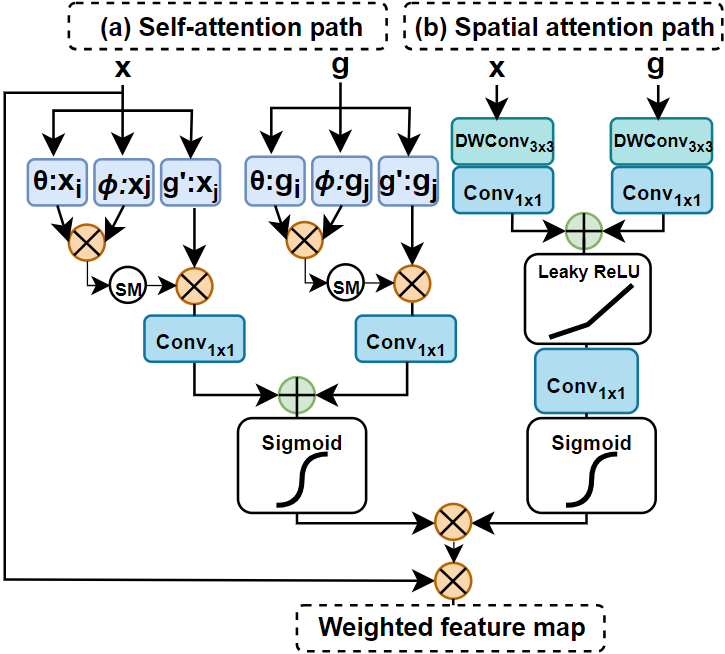}
    \caption{Dual attention gates.}
    \label{fig:AG}
\end{figure}
DAGs, illustrated in Fig. \ref{fig:AG}, consist of two paths: self-attention (a) and spatial attention (b).
In the self-attention path we adopt a type of attention similar to the one used in transformers, in which each position in the feature map $x_{i}$ is influenced by all other positions $x_{j}$, with the goal of effectively capturing global dependencies.\\
Let $\mathcal{X} \in R^{H\times W\times C_{x}}$ denote the skip connection feature map, where $H$, $W$, and $C_{x}$ represent height, width, and number of channels respectively. Let $x_{i}$ denote a vector along the third dimension (i.e. the channels dimension) of the tensor $\mathcal{X}$. Drawing inspiration from the principle of non-local processing in \cite{wang2018non}, the self-attention is mathematically expressed as follows:
\begin{equation}\text{A}(x_i) = \sum_{ j} \left( \frac{\exp{((W_{\theta} x_i)^T \cdot (W_{\phi} x_j))}}{\sum_{k} \exp{((W_{\theta} x_i)^T \cdot (W_{\phi} x_k))}} \right)W_{g'} x_j,\end{equation}
where $\theta$, $\phi$, and $g'$ are three transformations, $i$ specifies the output position, and $j$ and $k$  iterate over all positions in the feature map. This formula computes the output at $i$ as a weighted sum of all features  $x_{j}$, with weights derived from the exponential similarity between $x_{i}$ and each $x_{j}$, normalized across all positions $k$. The result integrates global contextual information into each local feature $x_{i}$, enhancing the overall representation. 
This operation is applied to both the feature map $x$ coming from the skip connection and the gating signal $g$ coming from the previous layer. Subsequently, the results undergo $1\times1$ convolutions, are summed, and passed through a sigmoid activation to yield attention weights ranging from 0 to 1.\\ 
We further investigated the mechanism's efficiency and flexibility by experimenting with the self-attention formula. We explored variations by implementing shared weight matrices \((W_\theta\), \(W_\phi\), and \(W_g'\)), using the same weights across all three transformations, and a weightless approach that computed the similarity between \(x_i\) and \(x_j\) directly, without any transformations. These modifications were aimed at assessing their impact on the model's performance and its ability to focus attention effectively, the results are discussed in the subsequent sections.\\
On the other hand, in the spatial attention path, within the spatial attention path, we initially subject both $x$ and $g$ to Depthwise (DW) convolutions followed by 1×1 convolutions. The choice of the DW convolution is motivated by the goal of enhancing both channel and spatial information processing, as this kind of convolutions are tailored to process each input channel separately which allows for a more nuanced understanding of the spatial features within each channel. A subsequent LeakyReLU activation ensures non-linearity and gradient flow, followed by Sigmoid-activation, acting as a spatial mask to highlight key areas and produce attention weights ranging from 0 to 1.\\
The final attention weights are computed by multiplying the outputs of both the self-attention and spatial attention paths. This fusion strategy effectively transmits low-level details from the skip connections to higher-level representations.
\vspace{-0.5cm}

\subsection{Efficient positional encoding-free Transformer}
\label{ssec:trans}

While CNNs excel in local feature extraction, capturing essential global representations for accurate diagnosis remains challenging. To address this limitation, we integrate an Efficient Transformer into our architecture.  For optimal efficiency we use one transformer block which we chose to integrate in the bottleneck. The bottleneck, being placed directly after the encoder receives the most abstract and compressed representations of the input data. As a result, a transformer at this stage can capture global dependencies across the entire image. Another key for this positioning choice is efficiency, since in this level of the architecture, the spatial dimensions are smaller, and thus the number of calculations required by the self-attention mechanism is reduced.
Besides, we chose to omit positional encodings, taking into account that positional information is inherently maintained by the convolutional layers due to their weight sharing and local receptive fields, this spatial structure is preserved up to the bottleneck where the transformer block is placed.  Furthermore, skip connections help to reintroduce detailed spatial information from the encoder to the decoder.
Moreover, the Self-attention mechanism can learn relative positions through the context of surrounding tokens or features. In that respect, Haviv, Adi, et al. \cite{haviv2022transformer} carried out a comparative study between transformers with and without explicit positional encoding, and proved that transformers trained without positional encoding are still competitive with standard models. 
By adopting this approach, we harness the robust contextual processing abilities of the transformer while avoiding their high computational costs.

\vspace{-0.25cm}

\subsection{Loss functions}
\label{ssec:loss}

A significant challenge with medical images is the issue of class imbalance. This arises because of the uneven distribution of classes, with tumors or inflamed regions often being localized in specific areas of the image, leading to dominance of the background class. Such an imbalance can skew the learning process, as the model becomes biased towards the more prevalent background class, potentially compromising its ability to accurately identify and segment the less represented, yet clinically crucial regions.
To address this, we use a combination of Dice loss and Sparse Categorical Crossentropy (SCCE) loss. Dice loss handles class imbalance by focusing on the overlap between predictions and ground truth, prioritizing smaller yet critical areas. SCCE loss enhances per-pixel classification accuracy, vital for detailed segmentation. The combined loss is a weighted sum: \begin{equation}
    L_{\text{combined}} = w_{\text{Dice}} L_{\text{Dice}} + w_{\text{SCCE}}L_{\text{SCCE}}
\end{equation}
\section{EXPERIMENTS and RESULTS}
\label{sec:ex}
\subsection{Dataset and Implementation Details}
\label{ssec:data}

\textbf{Dataset:} To address the lack of publicly available annotated bladder cancer datasets, we collaborated with doctors from the Urology Department at the Mohammed V Military Hospital to develop a comprehensive dataset from scratch. The data collection process began with the recording of cystoscopy and Trans Urethral Resection of Bladder Tumors (TURBT) procedures. When a tumor is detected, the doctors perform a biopsy, and the extracted tissue is then sent to  pathological anatomy for confirmation. The medical experts, identified and marked regions of interest in each frame, corresponding to three categories: Tumor, Inflammation, and Cystite. A key point we focused on is diversity, making sure to include images captured using different cystoscopes, accommodating the variability in imaging technology. We also made sure to capture the difference in tumor appearance among patients. This diversity is critical for developing a robust model capable of generalizing across various real-world scenarios and patient populations.
The process of data collection is still ongoing, aiming to continuously enrich the dataset's diversity and volume. So far, we have accumulated 657 annotated frames, with $70\%$ used for training, $10\%$ for validation and $20\%$ for testing.
\begin{figure}[]
    \centering
    \includegraphics[width=6
    cm]{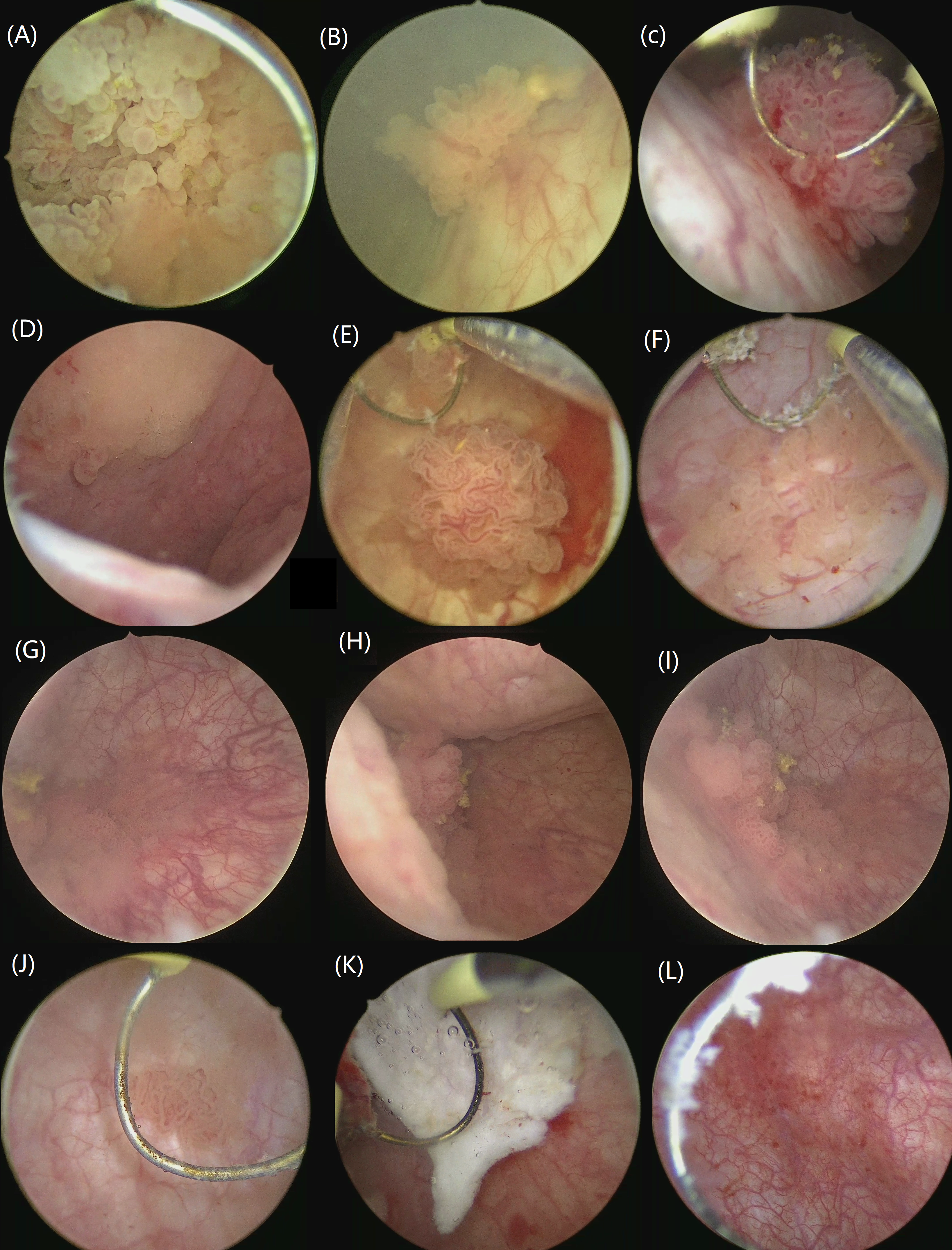}
    \caption{Cystoscopy data sample.}
    \label{fig:sample}
\end{figure}

Fig. \ref{fig:sample}  displays dataset samples, showcasing the diversity in tumor appearances. For example, image (A) depicts a tumor that occupies the entire frame, demonstrating cases where the tumor is extensively spread. In contrast, (B) and (E), show tumors that are centered in the middle. 
Challenging cases are highlighted in images such as (G), where tumors borders are obscured by surrounding inflammation, making it difficult to distinguish its edges. This complexity is critical for training models to recognize and segment tumors accurately under varying conditions. Similarly, smaller tumors like (J), which blend seamlessly into the background or nestle in the bladder's corners as seen in (H), represent another challenge due to their subtle presentation.
Additionally, the dataset includes images of cystites, such as the one shown in (K), where the inflammation's distinct appearance provides a contrast to tumor images. Image (L) further demonstrates an inflamed area, adding to the dataset's comprehensiveness by including various pathological presentations within the bladder.\\
\textbf{Implementation details:} Our model is trained with an Adam optimizer, a learning rate of \(10^{-3}\) and a batch size of 16. We employed a learning rate scheduler, reducing the learning rate by a factor of 0.1 when no improvement is observed.
Data augmentation techniques such as contrast adjustment, random rotation and scaling were used for improved generalization.\\
\textbf{Attention heads:} The main goal of this study is to develop a highly efficient, lightweight model that still delivers robust performance. To achieve this we have significantly reduced the model's size by decreasing the number of convolutional blocks, using DW separable convolutions and omitting positional encoding. However, a factor that accounts for a substantial parameter count is the multi-head attention. To address this, we conducted a comparative analysis of various attention head configurations, considering their impact on IoU, parameter count, and GFLOPs. The results in Table \ref{tab:att_heads} indicate that both the number of parameters and GFLOPs increase as the number of attention heads rises. We notice that beyond 4 attention heads, the IoU stabilizes at around $85\%$, while the computational complexity continues to grow. This trend indicates that the additional computational complexity introduced by more than 4 attention heads does not translate into significant gains in IoU, making 4 attention heads the optimal choice for balancing performance, parameter count, and computational efficiency.\\
\begin{table}[]
\centering
\begin{tabular}{l|c c c c}
\hline
\textbf{Number of att heads} & \textbf{\# params } & \textbf{GFLOPs} & \textbf{IoU$(\%)$ } \\
\hline
1 & 169474  & 2.6447600 & 81.50 \\ 
2 & 235394 & 2.6582913 & 83.51 \\  
3 & 301314 &  2.6718226 &  84.02\\  
\textbf{4} & \textbf{367234} & \textbf{2.6853540} & \textbf{85.70} \\ 
5 &433154 & 2.6988853 & 85.39 \\ 
6 & 499074 & 2.7124166 &  85.56 \\
7 & 564994  & 2.7259479 & 85.69 \\
8 & 630914 & 2.7394792 &  84.98\\
9 & 696834  & 2.7530106 &  85.73\\
10 & 762754 & 2.7665419 & 85.71 \\
\hline
\end{tabular}
\caption{Comparison of attention heads configurations}
\label{tab:att_heads}
\end{table}
\textbf{Self-attention:} To compute self-attention within the attention gates, experiments were performed to find the optimal similarity function. Initially, feature maps were projected using three transformations $\theta$, $\phi$, and $g'$, as in equation (1). Later experiments explored weight sharing among the three weight matrices $W_{\theta}$, $W_{\phi}$, $W_{g'}$, and even direct dot product similarity without weight projections.
\begin{figure}[] 
    \centering
    \includegraphics[width=9cm]{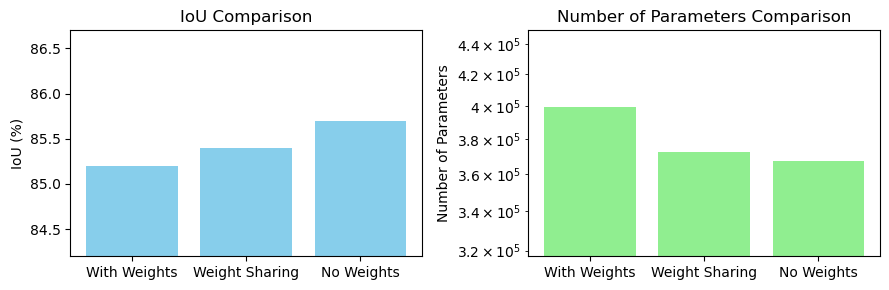}
    \caption{IoU and Parameters Against Weight Configurations.}
    \label{fig:weights}
\end{figure}
Figure \ref{fig:weights} displays the experimantal results. It is clear that adding weight matrices increases the parameters without a corresponding IoU improvement.  We chose to eliminate these extra weight matrices, opting for direct dot product similarity. This simplified the model, reducing parameters while maintaining or even improving IoU.\\
\textbf{Dual attention gates} We examine the impact of different attention mechanisms within the attention gates. Table. \ref{tab:attention_results} displays the IoU and parameters with attention gates utilizing solely spatial attention, self-attention, and a combination of both (DAG).
\begin{table}[ht]
\centering
\begin{tabular}{lccc}
\hline
\textbf{Attention Mechanism} & \textbf{IoU (\%)} & \textbf{\# params} & \textbf{GFLOPs} \\
\hline
Spatial Attention & 84.2 & 350882 & 1.992341 \\ 
Self Attention & 84.7 & 346511 & 1.853254\\ 
Dual Attention & 85.7 & 367234 & 2.685354 \\ 
\hline
\end{tabular}
\caption{Comparison of IoU scores and parameter counts for different attention mechanisms within the attention gates.}
\label{tab:attention_results}
\end{table}
The results in Table \ref{tab:attention_results} highlight dual attention as the most effective, achieving the highest IoU of $85.7\%$ with a moderate parameter count, demonstrating its superior capability to balance local and global contextual information.\\
\textbf{Loss function:} In order to mitigate the challenge of class imbalance, we employed a combined loss function. Experiments involving various weight configurations revealed that the optimal weights for the Dice loss and SCCE loss were determined to be \(w_{\text{Dice}} = 0.7\) and \(w_{\text{SCCE}} = 0.3\), respectively. This weighting strategy was found to be effective in addressing the issue of class imbalance within the scope of our experiments.
\begin{table*}[!ht]
\centering
\begin{tabular}{l|c c c c c }
\hline
\textbf{Method} & \textbf{\# params} & \textbf{GFLOPs} & \textbf{IoU  $(\%)$ } & \textbf{Difference}\\
\hline
Baseline & 32815 & 0.966405126 & 82.23 & - \\
Baseline + DAG & 69890 & 2.491517286 & 84.48 & +2.25 \\
Baseline + Transformer &330159  &  1.160241846 &  84.03 & +1.8\\
\textbf{Baseline + Transformer + DAG} & \textbf{367234} & \textbf{2.685354006} & \textbf{85.70} &\textbf{+3.47}\\
\hline
\end{tabular}
\caption{Ablation study for the proposed modules.}
\label{tab:ablation}
\end{table*}
\begin{table*}[!ht]
\centering
\begin{tabular}{l|c c c c c c c}
\hline
\textbf{Model} & \textbf{Dice $(\%)$ } & \textbf{IoU  $(\%)$ } & \textbf{ACC  $(\%)$ } & \textbf{Precision $(\%)$} & \textbf{Recall $(\%)$} & \textbf{\# params (M)}  \\
\hline
UNet \cite{ronneberger2015u}& 89.6 & 81.7 & 95.7 & 89.7   &  90.1 & 7.78 \\
Dilated UNet \cite{piao2019accuracy} & 91.0 & 83.8 & 96.1 & 90.4 & 91.9 & 7.78 \\
Attention UNet \cite{oktay2018attention} & 91.2 & 84.4 & 96.5 & 91.2 & \textbf{94.3} & 8.04 \\
TransUNet \cite{chen2021transunet} & \textbf{92.7} & \textbf{87.6} & \textbf{98.4} & \textbf{93.5} & 92.8 & 30.15  \\
Segformer-B0 \cite{xie2021segformer} & 91.8 &  84.6 & 96.4 & 92.3 & 91.4 &  3.71 \\
Segformer-B1 \cite{xie2021segformer} & 92.1 & 85.3  & 97.2 & 92.6 & 93.2&  13.67 \\
\textbf{Ours} & 92.0  & 85.7 & 96.9 & 92.2 & 92.1  & \textbf{0.36}  \\
\hline
\end{tabular}
\caption{Comparison with UNet, Dilated UNet, Attention UNet, TransUNet and Segformer B0 and B1.}
\label{tab:model_comparison}
\end{table*}
\begin{figure*}[!ht]
\centering
\includegraphics[width=12cm]{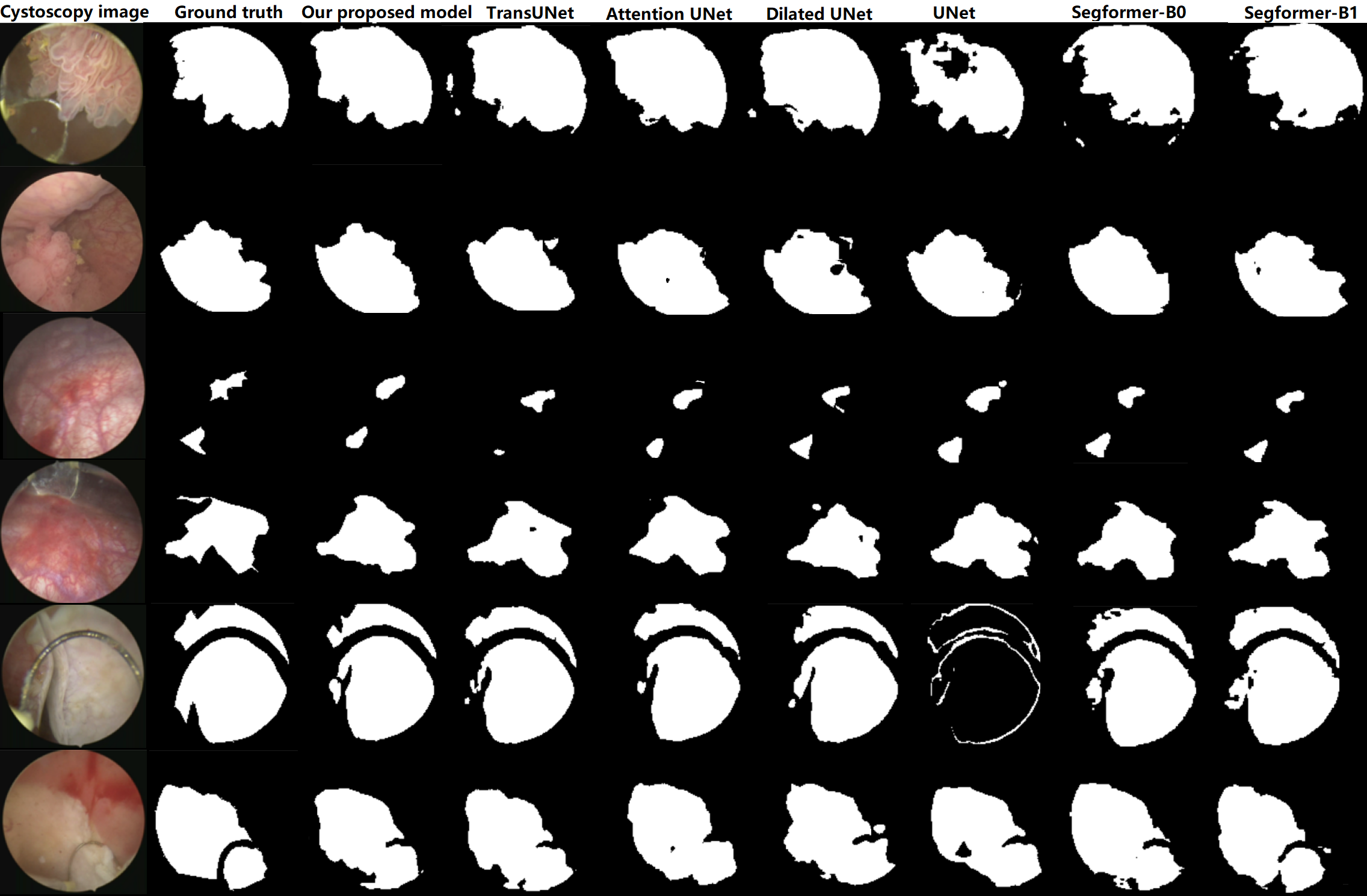}
\caption{Visual comparison with UNet, Dilated UNet, Attention UNet, TransUNet and Segformer-(B0,B1) on cystoscopy images.}
\label{fig:qualitative}
\end{figure*}
\vspace{-0.25cm}
\subsection{Ablation study}
\label{ssec:abl}
In order to validate the effectiveness of each component in our architecture, we conducted a series of ablation studies. The baseline model includes the encoder and decoder with DW separable convolutions.
The results in Table \ref{tab:ablation}, illustrate the impact of integrating DAGs and the transformer bottleneck. Initially, the baseline model achieves an IoU of $82.23\%$. The introduction of DAGs enhances this to $84.48\%$, affirming their effectiveness in boosting performance. Incorporating the transformer further improves the IoU to $84.03\%$, illustrating its capability in capturing complex data relationships. The most significant enhancement occurs when combining both DAGs and the transformer, which elevates the IoU to $85.70\%$. This marked improvement in IoU comes with an increase in parameters and GFLOPs. However, this is not a major concern, as the total parameters remain below one million and the GFLOPs are still very low, maintaining the model's overall efficiency and practicality for real-world applications.

\vspace{-0.25cm}

\subsection{Results and Comparison}
\label{ssec:res}
After training on our Cystoscopy dataset, our model achieved an accuracy of $96.9\%$, IoU of $85.7\%$, and a DC of $92.0\%$, all while maintaining a compact size with only $367234$ parameters. It outperforms UNet, Dilated UNet, and Attention UNet in terms of DC, IoU, and accuracy. Our model achieves competitive performance, even compared to TransUNet, which has significantly more parameters. Additionally, when considering lightweight architectures like Segformer B0 and B1, our model achieves better results despite having fewer parameters, making it suitable for resource-constrained scenarios. The comparative results are summarized in Table \ref{tab:model_comparison}. In summary, our strategic architectural decisions, allowed us to achieve a good performance while maintaining a highly efficient and lightweight model. This demonstrates the effectiveness of our approach in achieving a balance between model size and segmentation accuracy, even when compared to the lightest versions of Segformer, B0 and B1.\\
Fig. \ref{fig:qualitative} provides a comprehensive visual comparison of segmentation results. Upon visual examination, it becomes readily apparent that our model demonstrates a good precision in predicting boundaries. The visual analysis clearly reveals that our model's segmentation contours exhibit great accuracy and align closely with the true boundaries.

\vspace{-0.25cm}

\section{Conclusion}
\label{sec:con}
In conclusion, our study presents a hybrid segmentation model that excels in both accuracy and computational efficiency. Trained on Cystoscopy images from the Mohammed V Military Training Hospital, our model achieved excellent results with an accuracy of $96\%$, a mean IoU of $85\%$, and a Dice coefficient of $92\%$, all while maintaining a remarkably low parameter count of only 0.36 million. These results are not only a testament  to the model's effectiveness but also highlight its efficiency, crucial for real-time bladder cancer diagnosis.
 Despite its compact size, our model strongly rivals larger models in performance. This balance of high accuracy with a significantly reduced computational footprint positions our model as an attractive solution for medical scenarios where efficiency is as vital as accuracy.

{\small \bibliographystyle{IEEEbib}
\bibliography{strings,refs}}
\end{document}